\documentclass[letterpaper, 10 pt, conference]{ieeeconf}  

\IEEEoverridecommandlockouts                              

\usepackage{amsmath,amsfonts}
\usepackage{algorithm}
\usepackage{array}
\usepackage{amssymb}
\usepackage{graphicx}
\usepackage{cite}
\usepackage{float}
\usepackage{multirow}
\usepackage{algpseudocode}
\usepackage{booktabs}
\usepackage{wrapfig}
\usepackage[table]{xcolor}
\usepackage{makecell}
\usepackage{lipsum}
\usepackage{tabularx}
\usepackage{hyperref}

\title{\LARGE \bf
AION: Aerial Indoor Object-Goal Navigation Using Dual-Policy Reinforcement Learning
}

\author{Zichen Yan, Yuchen Hou, Shenao Wang, Yichao Gao, Rui Huang, Lin Zhao
\thanks{The authors are with the Department of Electrical and Computer Engineering, National University of Singapore, Singapore. {Email: \tt \small{zichenyan@u.nus.edu, zhaolin@nus.edu.sg}.} 
This work was supported by the Singapore Ministry of Education Tier 2 Academic Research Fund 
T2EP20224-0035. Corresponding author: Lin Zhao.
}
}

\begin{document}

\maketitle
\thispagestyle{empty}
\pagestyle{empty}

\begin{abstract}
  Object-Goal Navigation (ObjectNav) requires an agent to autonomously explore an unknown environment and navigate toward target objects specified by a semantic label. While prior work has primarily studied zero-shot ObjectNav under 2D locomotion, extending it to aerial platforms with 3D locomotion capability remains underexplored. 
  Aerial robots offer superior maneuverability and search efficiency, but also introduce new challenges in spatial perception, dynamic control, and safety assurance.
  In this paper, we propose AION for vision-based aerial ObjectNav without relying on external localization or global maps. AION is an end-to-end dual-policy reinforcement learning (RL) framework that decouples exploration and goal-reaching behaviors into two specialized policies. 
  We evaluate AION on the AI2-THOR benchmark and further assess its real-time performance in IsaacSim using high-fidelity drone models. Experimental results show that AION achieves superior performance across comprehensive evaluation metrics in exploration, navigation efficiency, and safety. The project is available at \url{https://github.com/Zichen-Yan/AION}.
\end{abstract}

\section{INTRODUCTION}
Enabling robots to find objects specified by text instructions is a fundamental capability for embodied AI. As a branch of embodied navigation, ObjectNav holds significant practical value across a wide range of applications, including service robots, disaster rescue, and autonomous inspection. Furthermore, zero-shot ObjectNav requires the agent to search for objects unseen during training. This task can be decomposed into two stages: exploration before the target is observed and goal-reaching after target detection. 

Indoor ObjectNav has been well developed on ground robots. Nevertheless, they typically rely on global localization and semantic maps for exploration~\cite{gervet2023navigating}. Given such maps, large language models (LLMs) demonstrate a strong potential for ObjectNav by enabling high-level scene reasoning and subgoal selection~\cite{zhang2025apexnav, zhou2023esc, yu2023l3mvn}.
However, indoor localization and mapping are prone to accumulated errors and can become unreliable in purely visual servoing systems. To avoid these issues and simplify manual fine-tuning across multiple modules, end-to-end methods~\cite{mjo,sc,zson,savn,tdanet} have emerged as a promising alternative, directly inferring actions from multi-modal observations.

In the aforementioned methods, agents are restricted to 2D locomotion with cameras typically mounted at a fixed altitude, making it difficult to search cluttered environments and find objects at different height levels. In contrast, aerial robots provide greater flexibility through 3D motion, allowing them to fly over obstacles and gain a broader aerial view. However, the expanded locomotion space also introduces new challenges. For RL policies, the 3D action space significantly increases training complexity and poses higher demands on spatial perception. Moreover, indoor map-based methods rely on accurate global poses and incur high 3D computational costs.
Beyond perception and localization, drone navigation is prone to collisions in cluttered scenes, which is often overlooked by prior work. Many algorithms are evaluated only in Habitat~\cite{DBLP:conf/iclr/PuigUSCYPDCHMVG24} or AI2THOR~\cite{ai2thor}, where motion execution is simplified by teleportation instead of dynamic control. The real-time performance and navigation safety under realistic robot dynamics remain unexplored. 
\begin{figure}[!t]
    \centering
    \includegraphics[width=\linewidth]{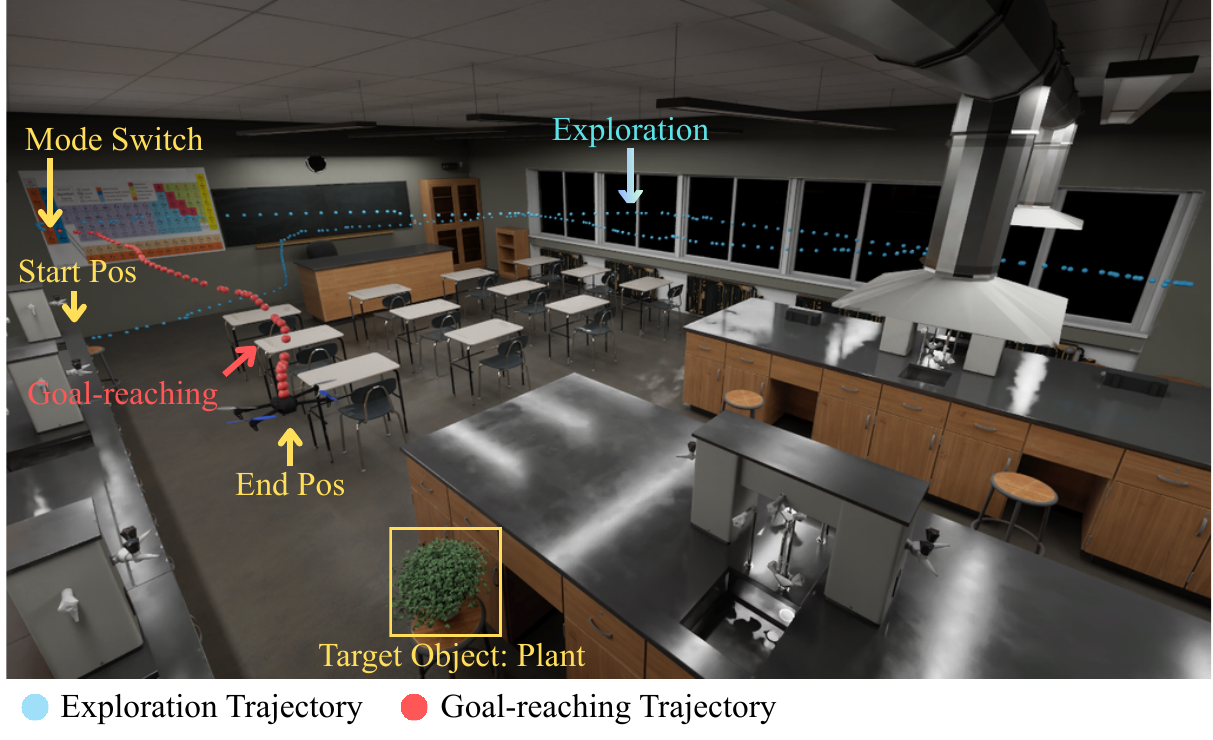}
    \caption{Demonstration of two-stage aerial ObjectNav with policy switching between exploration (AION-e) and goal-reaching (AION-g). The agent begins in exploration mode and switches to goal-reaching mode once the target object has been detected twice.}
    \label{fig:cover}
    \vspace{-1.5em}
\end{figure}

In this paper, we present \textbf{AION} (\textbf{A}erial \textbf{I}ndoor \textbf{O}bject-Goal \textbf{N}avigation), an end-to-end dual-policy RL framework designed to support 3D ObjectNav. As is shown in Fig.~\ref{fig:cover}, AION consists of two policies: AION-e for exploration, and AION-g for goal-reaching. The navigation performance is thoroughly evaluated on both the AI2THOR benchmark and the IsaacSim platform that simulates realistic physical dynamics. 
Our main contributions are as follows:
\begin{figure*}[!t]
    \centering
    \includegraphics[width=0.7\linewidth]{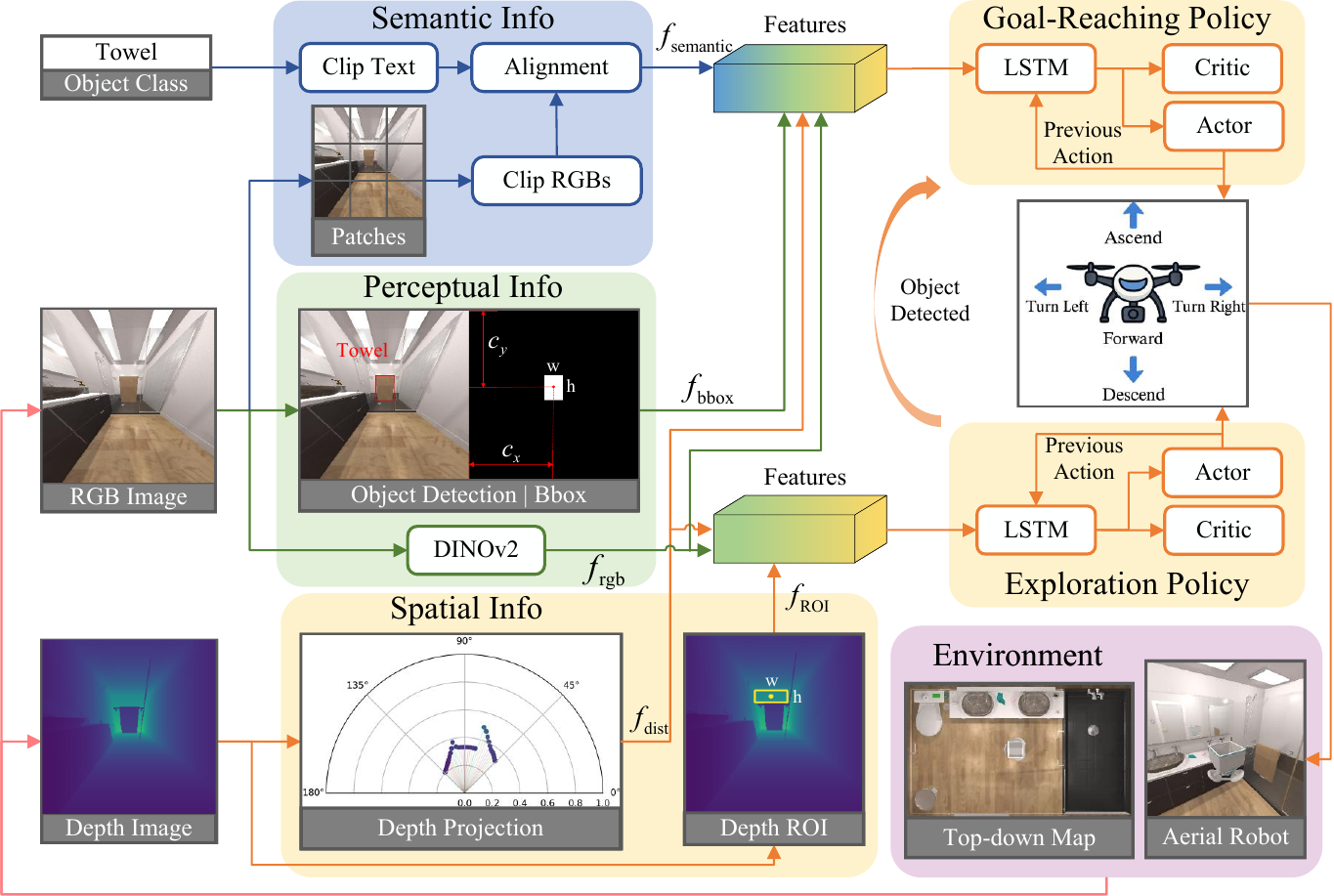}
    \caption{Overview of the proposed dual-policy RL framework for aerial indoor ObjectNav. RGB-D observations and textual classes are encoded into semantic, perceptual, and spatial features, which are fused as inputs to two policies: an exploration policy before target detection and a goal-reaching policy after detection.}
    \label{fig:framework}
    \vspace{-1em}
\end{figure*}

\begin{itemize}
    \item We extend indoor zero-shot ObjectNav from 2D to 3D locomotion. A dual-policy strategy is proposed to improve mapless exploration and goal-oriented navigation using RGB-D images, textual object classes, and altitude measurement. 

    \item Beyond improving benchmark metrics, AION emphasizes real-time performance and safety during aerial navigation, which is validated in IsaacSim under realistic drone dynamics.
\end{itemize}

\section{Related Works}
\subsection{Zero-Shot ObjectNav}
When perception and localization are fully known and accurate, modular methods have proven to be a mature solution for 2D zero-shot ObjectNav~\cite{gervet2023navigating}, which typically combines high-level planning and low-level goal-reaching while maintaining a semantic map for exploration. To accelerate the search, subgoal selection can be improved by rule-based methods~\cite{stubborn}, foundation models~\cite{yokoyama2024vlfm}, RL policies~\cite{chaplot2020object}, or powerful LLMs~\cite{yu2023l3mvn, zhou2023esc}. For pure visual servoing systems that lack LiDAR sensors, map-based methods are prone to unreliable indoor localization. To address this problem, Cai~\cite{cai2024bridging} explored mapless ObjectNav through scene reasoning by LLMs. 

Without explicit mapping, end-to-end methods aim to infer actions directly from multi-modal observations, or integrate latent localization and mapping into the planner~\cite{peng2025logoplanner}. 
Existing end-to-end methods focus on studying object relationship graphs~\cite{mjo, sc}, object-goal attention~\cite{tdanet, sun2024prioritized}, and memory mechanisms~\cite{du2023object}. To achieve zero-shot generalization, foundation models~\cite{clip,li2023blip} are utilized for vision-language grounding, allowing the agent to find novel object categories without requiring object-specific fine-tuning. DAT~\cite{dang2023search} is closely related to our work, as it adaptively fuses search and navigation modes within a unified framework. Different from DAT, we decompose the task into separate search and navigation modules to facilitate a fine-grained analysis, while further extending the formulation from 2D to 3D ObjectNav.

\subsection{Aerial ObjectNav on Drones}
Existing aerial ObjectNav research is largely explored under the vision-language navigation (VLN) paradigm, particularly in outdoor environments~\cite{lee2025citynav, xiao2025uav, liu2023aerialvln, wang2025uav}. 
Liu~\cite{liu2025indooruav} extended aerial VLN to indoor scenarios, but it was not validated on a drone system, where low-level control dynamics, real-time execution and navigation safety need to be considered. 
The nature of VLN on drones lies in grounding language and vision to follow human instructions. However, VLN struggles to explore the environment when prebuilt maps or detailed textual guidance are unavailable. Therefore, aerial VLN methods do not apply to classical ObjectNav tasks when only the target object class is specified. So far, 3D ObjectNav in indoor environments remains largely unresolved, and we focus on end-to-end vision-based solutions to avoid dependence on external localization and pre-built maps. 

\section{Proposed Methods}
As illustrated in Fig.~\ref{fig:framework}, the proposed framework employs an exploration policy AION-e while the object is undetected and switches to a goal-reaching policy AION-g once the object is detected. AION-e and AION-g adopt the same model architecture and use a partially shared set of multi-modal features. To enhance spatial understanding, we incorporate depth information into the end-to-end pipeline, allowing the agent to learn obstacle avoidance and efficient 3D navigation in cluttered environments.

\begin{figure*}[!t]
    \centering
    \includegraphics[width=1.0\textwidth]{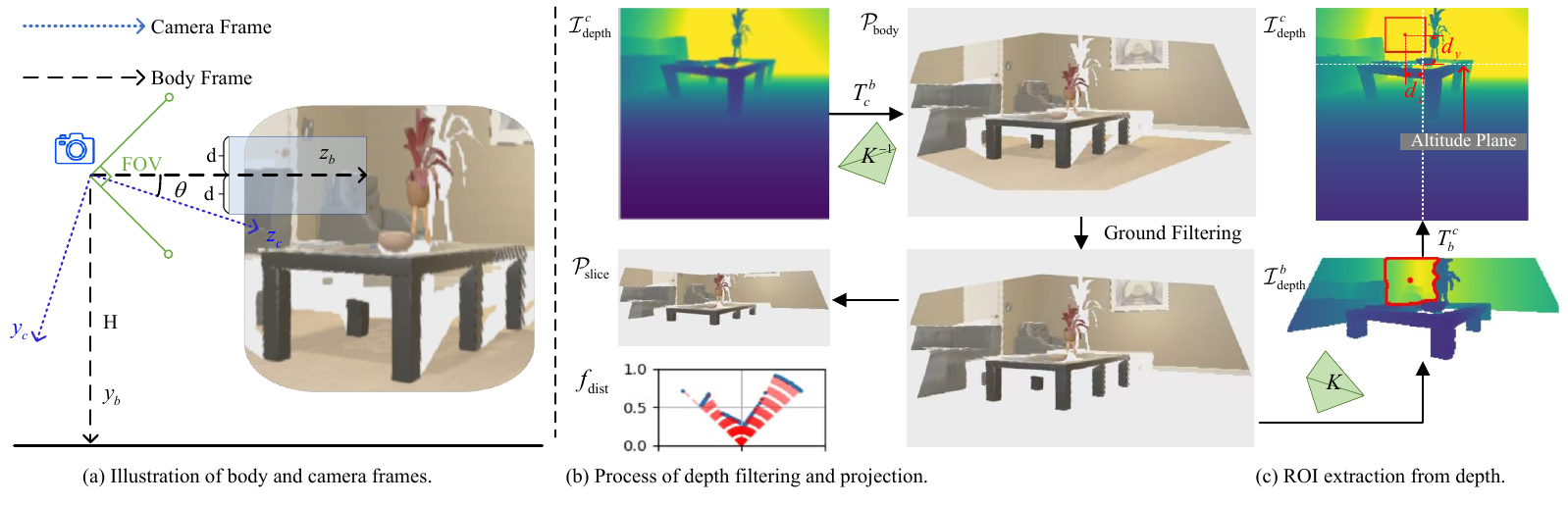}
    \caption{Process of spatial perception from depth images. (a) The geometry between the camera and body frames used to interpret RGB-D measurements. (b) Ground filtering and projection of depth into planar distance features. (c) Depth-based ROI extraction for exploration policy.}
    \label{fig:depth}
    \vspace{-1em}
\end{figure*}
\subsection{Task Definition} 
In an aerial ObjectNav task, the agent is supposed to find the target object $c_t$ from a class set $C$. 
The observations consist of an RGB image $\mathcal{I}_{\text{rgb}}\in\mathbb{R}^{H\times W \times3}$, a depth image $\mathcal{I}_{\text{depth}}\in\mathbb{R}^{H\times W}$, a target object class $c_t$ and the current altitude $H$.
The action space is extended to a 3D setting $\mathcal{A}=\{\textit{Ascend}, \textit{Descend},\textit{Forward}, \textit{TurnLeft}, \textit{TurnRight}, \textit{Done}\}$.
In the AI2THOR simulator, \textit{Ascend}, \textit{Descend}, and \textit{Forward} will move the aerial robot by $0.15~m$. \textit{TurnLeft} and \textit{TurnRight} will rotate the agent by $30^{\circ}$. The \textit{Done} action helps the agent learn to finish the task when the target object is close enough. The task is considered successful when 1) the bounding box center lies in the central $80\%$ of the image; 2) the distance to the goal is less than $1.5~m$; 3) the agent chooses \textit{Done}.

\subsection{Cross-Modality Attention}
To enable zero-shot understanding of textual goals, CLIP~\cite{clip} is used to ground vision with natural-language descriptions. Specifically, $\mathcal{I}_{\text{rgb}}$ is partitioned into $N$ patches $\{p_i\}_{i=1}^{N}$. Each patch is mapped to a visual embedding via the CLIP image encoder:
\begin{equation}
    v_i = \text{CLIP}_{\text{image}}(p_i) \in \mathbb{R}^d.
\end{equation}
Meanwhile, the CLIP text encoder generates the semantic embedding of the target object class $c_t$ via
\begin{equation}
    v_{c} = \text{CLIP}_{\text{text}}(c_t) \in \mathbb{R}^d.
\end{equation}
The cosine similarity for each patch-text pair is computed as
\begin{equation}
s_{i} = 
\frac{v_i^{\top}v_{c}}{
\|v_i\|_2 \, \|v_{c}\|_2} \in [-1,1], \ i=1,\ldots,N.
\end{equation}
The resulting similarity map $f_{\text{semantic}} = [s_i]_{i=1}^{N} \in \mathbb{R}^{N}$ indicates which visual regions are most relevant to the target.
The introduction of foundation models enables zero-shot ObjectNav by leveraging knowledge acquired from large-scale pretraining, without requiring any object-specific fine-tuning. Furthermore, the similarity map serves as an attention guide for the goal-reaching policy.
For perceptual information, we employ DINOv2~\cite{oquab2023dinov2} to extract visual features and an object detector such as YOLOv8~\cite{varghese2024yolov8} to obtain geometric cues from bounding boxes.

\subsection{Spatial Perception From Depth}
Depth information is essential for RL agents to perceive and reason about spatial structure. This section describes how depth images are leveraged to guide both obstacle avoidance and exploration, as is illustrated in Fig.~\ref{fig:depth}.

\subsubsection{Depth Projection for Obstacle Avoidance}
To enhance spatial perception for navigation safety, the raw depth $\mathcal{I}^c_{\text{depth}}$ in the camera frame is projected onto the horizontal plane at the current flight altitude to obtain a 2D laser scan $f_{\text{dist}}$, which indicates the closest distance to obstacles, as illustrated in Fig.~\ref{fig:depth}(b).
Specifically, given the camera intrinsic matrix $K$, pixels with homogeneous coordinate $\tilde{p}=[u,v,1]^\top$ and depth $z$ can be back-projected to a 3D point cloud in the camera frame as
\begin{equation}
    p_c=zK^{-1}\tilde{p}=[x_c,y_c,z_c]^\top.
\end{equation}
The point cloud $\mathcal{P}_{\text{body}}=\{p_b\}$ in the body frame can be obtained
by the homogeneous transformation of $\tilde{p}_c=[p_c^\top,1]^\top$:
\begin{equation}\label{eq:body}
\tilde{p}_b=T_{c}^b\tilde{p}_c=[p_b^\top,1]^\top=[x_b,y_b,z_b,1]^\top.
\end{equation}
where $T_{c}^b\in SE(3)$ is the extrinsic matrix from the camera frame to the body frame. We then apply a height filter on $\mathcal{P}_{\text{body}}$ to extract a horizontal slice after excluding the ground $\mathcal{P}_{\text{slice}}=\{p_b \in \mathcal{P}_{\text{body}} | -d \leq  y_b\leq d\}$, as shown in Fig.~\ref{fig:depth}(b). $\mathcal{P}_{\text{slice}}$ is then transformed from Cartesian to polar coordinates $\mathcal{P}_{\text{polar}}
= \{(\rho, \phi) \mid \rho \in [0,1], \phi \in [-\bar{\phi}, \bar{\phi}]\}$, where 
$\rho$ and $\phi$ denote the normalized radial distance and azimuth angle, respectively. $\bar{\phi}$ is determined by the camera's horizontal field of view. $\mathcal{P}_{\text{polar}}$ can be uniformly divided into $N$ sectors according to the range of $\phi$, i.e., $\mathcal{P}_{\text{polar}}= \bigcup_{k=1}^{N} \mathcal{P}_{\text{polar}}^{k}$, where $ \mathcal{P}_{\text{polar}}^{k}$ is defined as
\begin{equation}
\mathcal{P}_{\text{polar}}^{k}
= \{ (\rho,\phi) \in \mathcal{P}_{\text{polar}} \mid
\phi \in [-\bar{\phi} + (k-1)\tfrac{2\bar{\phi}}{N},
          -\bar{\phi} + k\tfrac{2\bar{\phi}}{N}) \}.
\end{equation}
For each sector $\mathcal{P}_{\text{polar}}^{k}$, the closest distance to obstacles is 
\begin{equation}\label{eq:dist}
\rho_k^{\min}=\min_{(\rho,\phi)\in \mathcal{P}_{\text{polar}}^{k}}\rho.
\end{equation}
Concatenating $\rho_k^{\min}$ from the $N$ sectors produces the 2D laser scan-like feature
$f_{\text{dist}}=[\rho_1^{\min}, \ldots, \rho_N^{\min}] \in \mathbb{R}^N$,
which indicates the distribution of nearby obstacles. 

\subsubsection{Depth ROI for Exploration}
Vision-based pose estimation is prone to drift, and a mapless exploration strategy can mitigate this issue by eliminating the need for explicit mapping. Inspired by \cite{sun2025frontiernet} that guides indoor exploration by visual cues, regions with larger depth values often indicate open and navigable space with increased probability of finding the target. 
To this end, the point cloud $\mathcal{P}_{\text{body}}$ from Eq.~\eqref{eq:body} is used to extract a Region of Interest (ROI) for visual tracking. As shown in Fig.~\ref{fig:depth}(c), the ground in $\mathcal{P}_{\text{body}}$ is first filtered, and the remaining points are projected onto the image plane using the camera intrinsics $K$ to obtain $\mathcal{I}^{b}_{\text{depth}}$. 
Then, the farthest valid regions are identified by thresholding the upper percentile of the depth values. Among these regions, the connected component with the largest area is selected as the ROI, from which the bounding box is extracted. Finally, the ROI feature $f_{\text{ROI}}$ is presented as
\begin{equation}\label{eq:ROI}
    f_{\text{ROI}} = [d_x,d_y,\bar{z}_{\text{ROI}},H]^{\top}.
\end{equation}
where $d_x,d_y \in [0,1]$ represent the normalized offset of the box centroid from the origin of the current altitude plane, as illustrated in Fig.~\ref{fig:depth}(c). $\bar{z}_{\text{ROI}}$ denotes the mean depth of the ROI in $\mathcal{I}^{b}_{\text{depth}}$. $H$ denotes the current altitude. In summary, $f_{\text{ROI}}$ captures directional information from the agent to the ROI target, which is essential for the exploration policy.

\subsection{Goal-Reaching Policy Training} 
On the iTHOR~\cite{ai2thor} benchmark, we extend the action space for 3D ObjectNav. Policies trained on iTHOR will be used solely for goal-reaching due to their limited exploration capability.
The class set $C=\{S, U\}$, where $S$ is defined as the seen class set during the policy training and $U$ denotes the unseen set.
In the training phase, the initial agent pose and the object class $c_t \in S$ are randomly selected at the beginning of each episode. In the testing phase, the target object class is chosen from $\{S,U\}$ to evaluate its generalization to unseen objects. 
Following previous works~\cite{zson, savn, tdanet}, RL training adopts parallel environments and Asynchronous Advantage Actor-Critic (A3C) to enable efficient asynchronous learning and improve generalization across diverse scenarios. 
The key challenge lies in designing appropriate multi-modal inputs and reward functions to effectively guide the policy behavior, as described below.
\subsubsection{Approaching Reward}
\begin{equation}
    R_d=\|d_{t-1}-d_t\|_2,
\end{equation}
where $d_t$ denotes the distance to the closest target object at time $t$.
\subsubsection{Parent Reward}
Following MJO~\cite{mjo}, the agent will receive a reward when it observes the target’s parent object for the first time, 
\begin{equation}
    R_{\text{parent}} = 
    \begin{cases}
        r_p, & \text{if target’s parent is seen}, \\
        0,  & \text{otherwise}.
    \end{cases}
\end{equation}
It helps reduce exploration difficulty by guiding the agent toward reasonable regions that increase the likelihood of locating the target.
\subsubsection{Bounding Box Reward}
The bounding box of the target object is obtained either from the AI2THOR simulator or existing object detection algorithms. For the perceptual module illustrated in Fig.~\ref{fig:framework}, the feature $f_{\text{bbox}}$ contains specific geometric cues of detected objects, defined as
\begin{equation}
    f_{\text{bbox}}=[c_x,c_y,w,h,S_{\text{bbox}}/S_{\text{all}},H]
\end{equation}
where $c_x,c_y$ are the normalized center coordinates of the bounding box, and $w,h$ represent its normalized width and height. $S_{\text{bbox}}/S_{\text{all}}$ indicates the relative area of the object within the image. $H$ corresponds to the current altitude.
The bounding box reward is designed as
\begin{equation}
    R_{\text{bbox}}=\min (S_{\text{bbox}}/S_{\text{all}}, 0.1),
\end{equation}
which encourages the agent to approach the target.
The final reward for goal-reaching is summarized as 
\begin{equation}\label{eq:rg}
r_g=R_d+R_{\text{parent}}+R_{\text{bbox}}+R_{\text{suc}}+R_c+\gamma,
\end{equation}
where $R_{\text{suc}}=+5$ is the success reward, $\gamma=-0.02$, $R_c=-0.1$ are the step and collision penalty, respectively.

\subsection{Exploration Policy Training} \label{sec:exp}
The iTHOR benchmark mainly includes single-room scenarios, where the agent can observe the target simply by flying higher and scanning the surroundings. Consequently, the trained goal-reaching policy exhibits limited exploration capability. To improve the exploration, we train an extra exploration policy on ProcTHOR~\cite{procthor}, a more complicated benchmark featuring multi-room layouts. The exploration policy aims to approach the depth ROI in Fig.~\ref{fig:depth}(c) while avoiding obstacles. The reward function that encourages exploration is designed as follows:
\subsubsection{Approaching Reward}
In Eq.~\eqref{eq:ROI}, the mean depth of the ROI $\bar{z}_{\text{ROI}}$ reflects whether the agent is moving closer to the ROI target. Thus, the approaching reward is defined as
\begin{equation}
R_{\text{forward}}=\min\left(\max\left(  \bar{z}_{\text{ROI}}^{t+1} - \bar{z}_{\text{ROI}}^{t}, -0.2\right), 0.2\right).
\end{equation}

\subsubsection{Direction Reward} To move toward the right direction, the agent must learn how to correct its pose through turning maneuvers and vertical motions. In this case, the coordinates of the ROI centroid $d_x,d_y$ in Eq.~\eqref{eq:ROI} provide crucial guidance for direction correction, leading to the direction reward
\begin{equation}
R_{\text{dir}} =
\begin{cases}
-0.75 d, & d > d_{\text{thr}}, \\
0, & d \le d_{\text{thr}}.
\end{cases}, d=\sqrt{d_x^{2} + (2d_y)^{2}}
\end{equation}
where the threshold $d_{\text{thr}}=0.3$ helps avoid oscillatory behavior caused by frequent direction correction. The coefficient of $d_y$ is set as 2 to enhance the sensitivity to height variations.
\subsubsection{Safety Reward} For aerial robots, safety issues are critical in cluttered indoor environments. Obstacle avoidance relies heavily on the feature $f_{\text{dist}}=[\rho_1^{\min}, \ldots, \rho_N^{\min}]$ from Eq.~\eqref{eq:dist}. We denote a safety distance by $\rho_{\min}=\min_{i=1:N} \rho_i^{\min}$. To encourage the agent to steer away from nearby obstacles, the safety reward is designed as 
\begin{equation}
R_{\text{safe}}=
\begin{cases}
1-e^{2(\rho_{\text{thr}} - \rho_{\min})}, & \rho_{\min} \le \rho_{\text{thr}}, \\
0, & \rho_{\min} > \rho_{\text{thr}},
\end{cases}
\end{equation}
where the safety threshold $\rho_{\text{thr}}={1}/{3}$ corresponds to a physical distance of $1~m$ in our experiments for a maximum depth range of $3~m$. Finally, the reward for the exploration policy training is summarized as 
\begin{equation}
r_e=R_{\text{forward}}+R_{\text{dir}}+R_{\text{safe}}+\gamma,
\end{equation}
where $\gamma=-0.01$ is the step penalty.
\section{Experiments}
In this section, we conduct experiments to analyze the impact of multi-modal inputs and different rewards on indoor aerial ObjectNav. All algorithms are trained on the AI2THOR platform for a fair comparison. Four metrics are used to evaluate navigation performance: Success Rate (SR), Success weighted by Path Length (SPL), Collision Rate (CR), and Free-space Coverage Ratio (FCR). FCR is computed on the top-down occupancy map as the ratio of observed free-space cells to all free-space cells.
\begin{table}[!t]
\centering
\caption{Details of Seen/Unseen (S/U) Object Splits.}
\label{Tab:1}
\renewcommand{\arraystretch}{1.25}
\begin{tabularx}{\columnwidth}{
>{\centering\arraybackslash}m{0.9cm}
>{\raggedright\arraybackslash}m{3.3cm}
>{\raggedright\arraybackslash}m{3.4cm}
}
\toprule
\textbf{Room} & \textbf{18/4 Split} & \textbf{14/8 Split} \\
\midrule

Kitchen
&
\textbf{S:} \texttt{Spatula}, \texttt{Bread}, \texttt{Mug}, \texttt{CoffeeMachine}, \texttt{Apple} \newline
\textbf{U:} \texttt{Toaster}
&
\textbf{S:} \texttt{Spatula}, \texttt{Bread}, \texttt{Mug}, \texttt{CoffeeMachine} \newline
\textbf{U:} \texttt{Apple}, \texttt{Toaster} \\
\midrule

\makecell{Living \\ Room}
&
\textbf{S:} \texttt{Painting}, \texttt{Vase}, \texttt{RemoteControl}, \texttt{ArmChair}, \texttt{Television} \newline
\textbf{U:} \texttt{Laptop}
&
\textbf{S:} \texttt{Painting}, \texttt{Vase}, \texttt{RemoteControl}, \texttt{ArmChair} \newline
\textbf{U:} \texttt{Television}, \texttt{Laptop} \\
\midrule

Bedroom
&
\textbf{S:} \texttt{Blinds}, \texttt{DeskLamp}, \texttt{Book}, \texttt{AlarmClock} \newline
\textbf{U:} \texttt{Pillow}
&
\textbf{S:} \texttt{Blinds}, \texttt{DeskLamp}, \texttt{Book} \newline
\textbf{U:} \texttt{AlarmClock}, \texttt{Pillow} \\
\midrule

Bathroom
&
\textbf{S:} \texttt{SoapBar}, \texttt{Towel}, \texttt{SprayBottle}, \texttt{Mirror} \newline
\textbf{U:} \texttt{ToiletPaper}
&
\textbf{S:} \texttt{SoapBar}, \texttt{Towel}, \texttt{SprayBottle} \newline
\textbf{U:} \texttt{ToiletPaper}, \texttt{Mirror} \\
\bottomrule
\end{tabularx}
\vspace{-1.5em}
\end{table}

\subsection{Training and Evaluation Setup}
The policy training needs online interaction with the AI2THOR simulator, where the iTHOR benchmark includes 120 realistic indoor scenarios, 80 rooms for training and 40 rooms for evaluation. 
The unseen/seen splits in Table~\ref{Tab:1} are provided by previous benchmarks~\cite{mjo, tdanet}. There are 4 different types of room layouts: Kitchen, Living room, Bedroom, and Bathroom. Each kind of room contains different target object classes, divided into the seen and unseen sets. The goal-reaching policy AION-g and ObjectNav baselines are trained for $1\times10^{7}$ steps on iTHOR using two RTX 4090 GPUs, and then evaluated on the same test set used in \cite{mjo}. Each room type is tested for 150 episodes. 

Since iTHOR can not provide sufficient scene complexity to learn exploration behaviors, we train the exploration policy AION-e for $4\times10^{6}$ steps on ProcTHOR, which contains over 10,000 procedurally generated indoor scenes with substantially larger spatial scale and layout diversity.

\subsection{Baseline Models for Goal-Reaching}
On the iTHOR benchmark, we compare AION-g with state-of-the-art (SOTA) and open-source baselines, which are originally designed for end-to-end 2D ObjectNav to handle both exploration and goal-reaching. 
To adapt these baselines to the proposed 3D setting, we keep their original network architectures unchanged and extend their action spaces to 3D navigation by adding vertical motions.
To ensure a fair comparison in terms of available information, we augment the feature embeddings of all baselines with the bounding box feature $f_{\text{bbox}}$, as reported in Table~\ref{tab:2}. All baselines are retrained from scratch in 3D ObjectNav tasks. 

\textbf{BaseModel}~\cite{savn} extracts visual features from $\mathcal{I}_{\text{rgb}}$ using CNNs and encodes the target object category using a GloVe~\cite{glove} embedding. All the feature embeddings are concatenated and fed into an LSTM-based actor-critic policy trained by A3C.

\textbf{Scene Prior}~\cite{sc} introduces scene-prior knowledge through Graph Convolutional Networks (GCNs), which model semantic relationships between object categories and scenes.

\textbf{MJO}~\cite{mjo} performs ObjectNav by exploiting hierarchical object relationships, such as the relation between a target object and its parent object. Instead of relying on visual backbones, it uses object detections, GCN-based object-relation reasoning, and GloVe embeddings for target representation.

\textbf{SSNet}~\cite{zson} is a zero-shot ObjectNav method that uses semantic attention to guide exploration toward objects that are semantically related to the target. It computes a similarity map based on GloVe embeddings and uses this semantic guidance for navigation. 
\subsection{Evaluation of Goal-Reaching Policy}

\begin{table*}[!t]
\centering
\caption{Comparison Results of Zero-Shot ObjectNav on the iTHOR benchmark.}
\renewcommand{\arraystretch}{1.15}
\begin{tabular}{lccccccc}
\toprule
\multicolumn{1}{c}{\multirow{2}{*}{\textbf{Model}}}&\multirow{2}{*}{
\makecell{\textbf{Seen/Unseen}\\\textbf{split}}} &
\multicolumn{3}{c}{\textbf{Seen class}} &
\multicolumn{3}{c}{\textbf{Unseen class}} \\
\cline { 3 - 8 }
& & SR(\%) $\uparrow$ & SPL(\%) $\uparrow$ & CR(\%) $\downarrow$ & SR(\%) $\uparrow$ & SPL(\%) $\uparrow$ & CR(\%) $\downarrow$ \\
\midrule
BaseModel~\cite{savn}(w/ $f_{\text{bbox}}$)     & \multirow{5}{*}{18/4} & 76.7 &  39.9 & 5.5  &  81.5  &  36.4& \textbf{2.3}\\
Scene Prior~\cite{sc}(w/ $f_{\text{bbox}}$)  &  & 74.3  &  42.1& \textbf{5.5} & 83.7  &  41.9 & 5.4 \\
MJO~\cite{mjo}(w/ $f_{\text{bbox}}$)   &  & 81.2 &  52.0& 11.4 & 90.7  & 51.7& 14.2 \\
SSNet~\cite{zson}(w/ $f_{\text{bbox}}$)    &  & 72.3 &  50.4& 9.8 & 77.8 & 50.0 & 12.2\\
\rowcolor{gray!20}\textbf{AION-g} & & \textbf{88.7} &  \textbf{57.9} & 5.6 &  \textbf{95.0} & \textbf{55.2}& 7.6 \\
\midrule
BaseModel (w/ $f_{\text{bbox}}$)    & \multirow{5}{*}{14/8} & 73.3 &  47.3 & 10.9  &  70.8  &  46.6& 9.1\\
Scene Prior (w/ $f_{\text{bbox}}$) & & 79.3 &  52.7 & 6.5  &  71.0  &  44.8& 5.1 \\
MJO (w/ $f_{\text{bbox}}$)  & & 78.8 &  43.6 & 10.9  &  83.0  &  45.6 & 11.7 \\
SSNet (w/ $f_{\text{bbox}}$)   & & 79.2 &  44.3 & 6.8& 81.8 & 46.4& \textbf{4.5} \\
\rowcolor{gray!20}\textbf{AION-g} & & \textbf{84.7} &  \textbf{61.2} & \textbf{4.8}  &  \textbf{87.0} &  \textbf{60.5} & 5.2 \\
\bottomrule
\end{tabular}
\label{tab:2}
\end{table*}

\begin{table}[!t]
\centering
\caption{Ablation Studies on Multi-modal inputs and reward designs.}
\setlength{\tabcolsep}{2pt}
\renewcommand{\arraystretch}{1.15} 
\begin{tabular}{c|lcccccc}
\toprule
\multicolumn{2}{c}{\multirow{2}{*}{\textbf{Model}}} &
\multicolumn{3}{c}{\textbf{18 Seen class}} &
\multicolumn{3}{c}{\textbf{4 Unseen class}} \\
\cline { 3 - 8 }
\multicolumn{2}{c}{} & SR $\uparrow$ & SPL $\uparrow$ & CR $\downarrow$& SR $\uparrow$ & SPL $\uparrow$ & CR $\downarrow$\\
\midrule
\multicolumn{2}{c}{AION-g} & 88.7 &  57.9& 5.6 &  95.0 & 55.2& 7.6\\
\cline{1-2}\cline{3-8}
\multirow{4}{*}{\rotatebox{90}{Inputs}} 
&w/o $f_{\text{semantic}}$ & 81.8 &  49.2 & 6.1  &  88.5  &  44.6& 5.8\\
&w/o $f_{\text{dist}}$ & 82.8 &  39.9 & 7.1& 88.3 &  36.7 & 7.0\\
& \cellcolor{gray!20}w/o $f_{\text{bbox}}$  & \cellcolor{gray!20}14.0 &  \cellcolor{gray!20}8.6 & \cellcolor{gray!20}1.1  &  \cellcolor{gray!20}16.2 & \cellcolor{gray!20}9.1 & \cellcolor{gray!20}1.5\\
&w/o $f_{\text{rgb}}$  & 77.0 &  52.6 & 8.0  &  80.3  &  51.5& 7.9\\
\cline{1-2}\cline{3-8}
\multirow{4}{*}{\rotatebox{90}{Reward}} 
&w/o $R_{\text{bbox}}$ &  83.3 &  48.0 & 5.1&  92.3 & 49.8 & 4.6\\
&w/o $R_{d}$ & 77.5 &  33.9 & 4.2  &  86.7  &  33.4& 3.5 \\
&w/o $R_{\text{parent}}$  & 86.5 & 58.3 & 7.3  & 96.7 & 62.3& 7.1\\
&\cellcolor{gray!20}w/o $R_c$ & \cellcolor{gray!20}83.8 &  \cellcolor{gray!20}39.4 & \cellcolor{gray!20}26.4  & \cellcolor{gray!20}93.5 & \cellcolor{gray!20}35.9& \cellcolor{gray!20}34.0\\
\bottomrule
\end{tabular}
\label{tab:3}
\end{table}

\begin{figure}[!t]
    \vspace{-1.5em}
    \centering
    \includegraphics[width=1\linewidth]{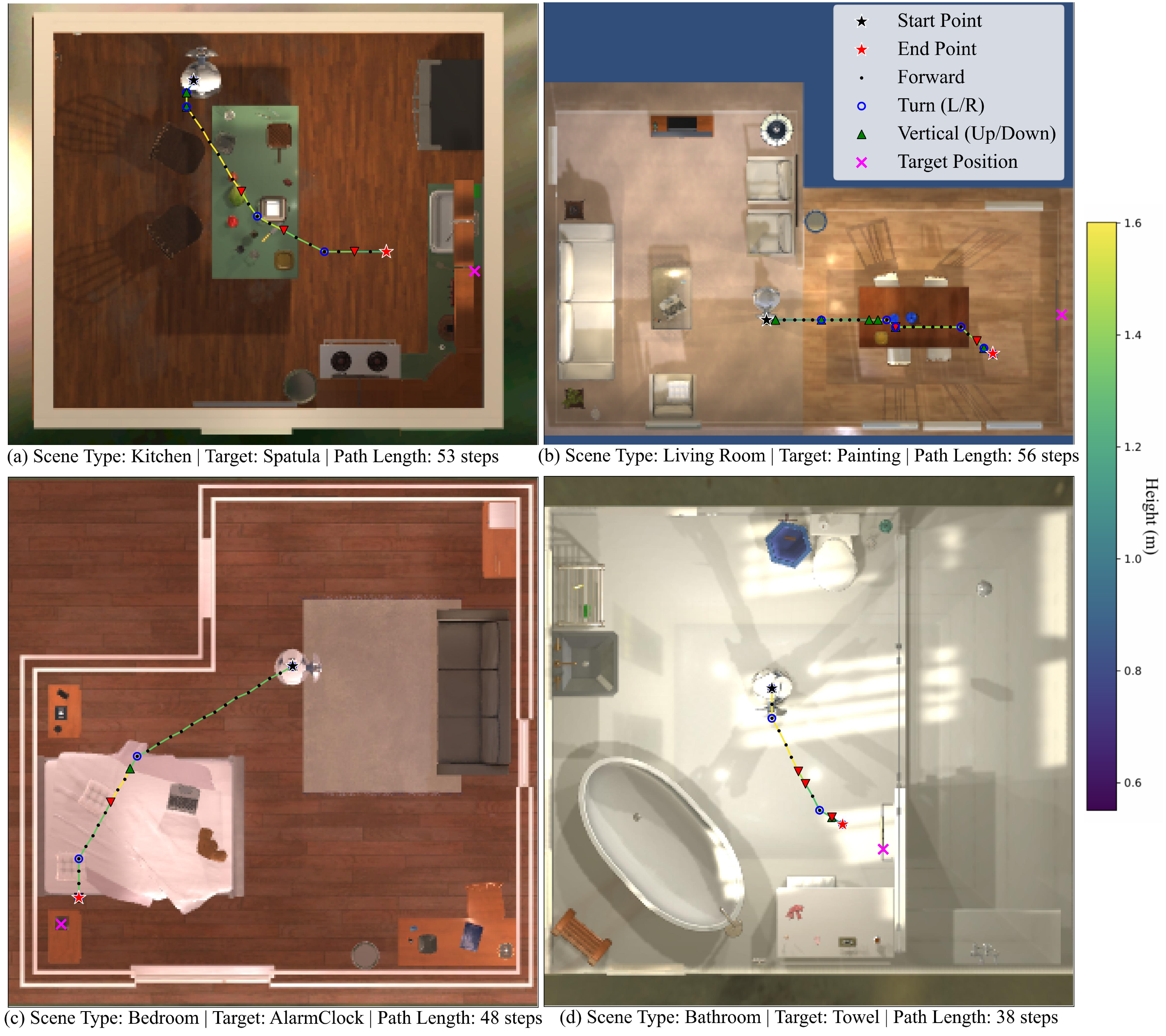}
    \caption{ObjectNav trajectories of AION-g across different kinds of iTHOR rooms, including Kitchen, Living Room, Bedroom, and Bathroom.}
    \label{fig:room}
    \vspace{-1.5em}
\end{figure}

Table~\ref{tab:2} presents the comparison results of goal-reaching performance. 
For both seen and unseen objects, AION-g achieves the best performance in SR and SPL while maintaining a relatively low CR. Among the baselines, MJO achieves almost the highest SR. An increase in SR often leads to a higher CR, as the agent tends to take greater risks to approach the goal aggressively. In contrast, AION-g outperforms MJO by (\textbf{+7.5\%/+4.3\%}) in SR and (\textbf{+5.9\%/+3.5\%}) in SPL on the 18/4 split, with a notably lower CR. 
A similar trend can be observed on the 14/8 split, where AION-g outperforms the baselines across nearly all metrics. Our model not only achieves strong generalization to unseen objects but also enhances safety, which is largely attributed to depth observations. 
The navigation trajectories of AION-g in four different types of rooms are visualized in Fig.~\ref{fig:room}, where the chosen action at each step is annotated.

\subsection{Ablation Study on Goal-Reaching Policy}
To further analyze the contributions of different rewards and input features to the goal-reaching performance, ablation experiments are conducted and summarized in Table~\ref{tab:3}. Among the multi-modal inputs, removing $f_{\text{bbox}}$ causes a dramatic drop in both SR and SPL, indicating that bounding box information is crucial for aerial ObjectNav. Regarding the reward design, canceling collision penalty $R_c$ leads to a sharp increase in CR. The parent reward $R_{\text{parent}}$ exhibits a limited contribution, as its removal does not noticeably affect performance metrics. Overall, the remaining components all contribute positively to navigation performance. 

\section{Evaluation on Drones in IsaacSim}
\begin{figure*}[!t]
    \centering
    \includegraphics[width=0.71\textwidth]{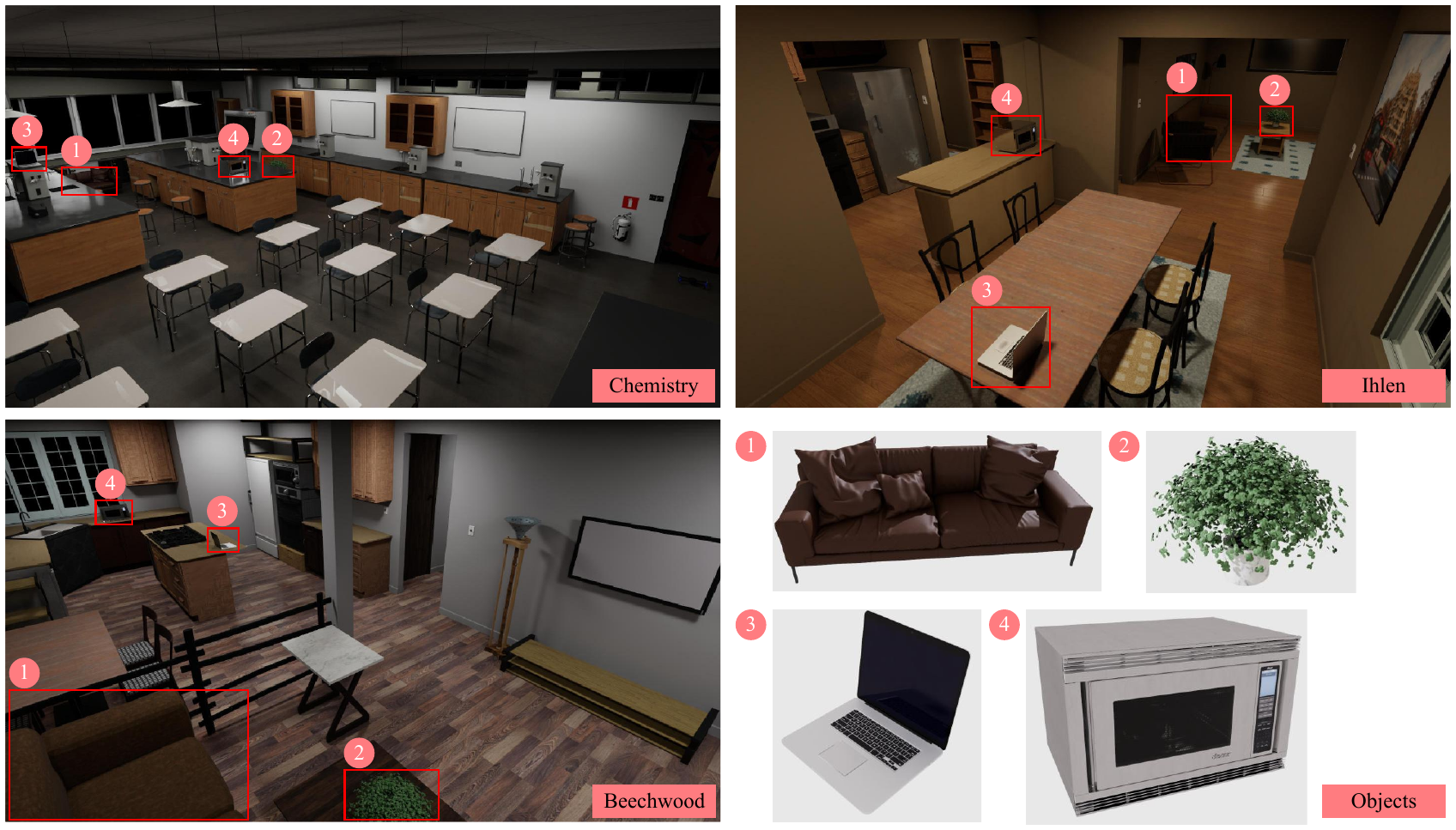}
    \caption{Placement of unseen objects at varying heights across different indoor environments in IsaacSim.}
    \label{fig:isaacsim}
    \vspace{-1em}
\end{figure*}

To test the real-time performance of algorithms under realistic drone dynamics, we conduct further experiments using IsaacSim~\cite{NVIDIA_Isaac_Sim} with Pegasus~\cite{Pegasus}.
IsaacSim is well-known for providing photo-realistic rendering and high-fidelity physics simulation, while Pegasus integrates the PX4 autopilot and ROS2, allowing it to closely approximate real-world flight. As shown in Fig.~\ref{fig:isaacsim}, scenes are provided by OmniGibson~\cite{omnigibson} with three houses selected for drone experiments: \textit{Chemistry}, \textit{Beechwood}, and \textit{Ihlen}. 

In our experiments, discrete actions from RL policies are interpreted as velocity references for the PX4 autopilot to track. Specifically, \textit{Ascend} and \textit{Descend} correspond to vertical velocities of $\pm0.15~m/s$ along the $Z$-axis; \textit{Forward} applies a forward velocity of $0.8~m/s$; and the angular velocity is $\pm 25~^\circ/s$ for \textit{TurnLeft} and \textit{TurnRight}. The inference frequency of AION policies reaches about 2.7 Hz on the RTX 4080S GPU. The computational cost mainly comes from model inference (YOLO, CLIP, DINO) and depth processing.
Considering the latency of control response, each RL action is executed for a duration of $0.5~s$ to avoid oscillations.

\subsection{Evaluation of Exploration Policy}
\begin{figure}[!t]
    \centering
    \includegraphics[width=0.75\linewidth]{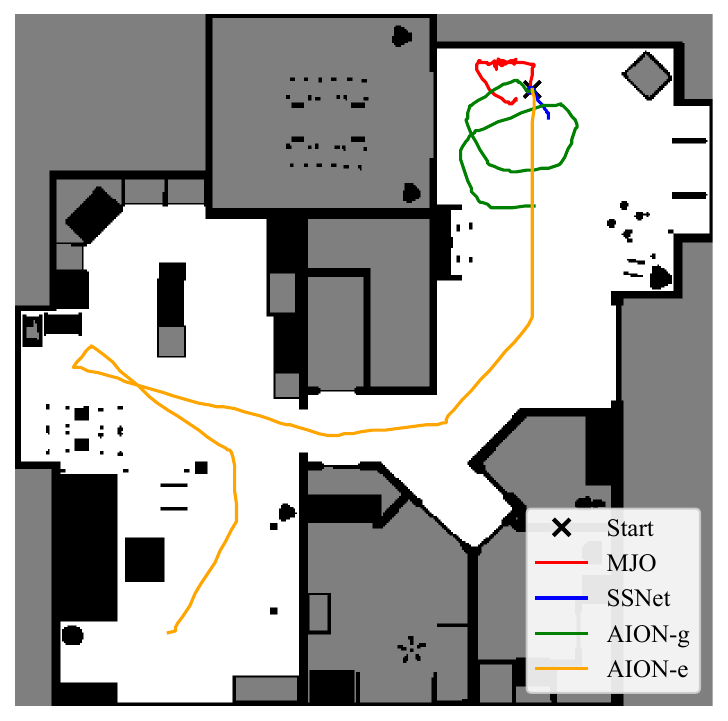}
    \caption{Top-down view of 3D exploration trajectories in \textit{Beechwood}.}
    \label{fig:exp}
    \vspace{-1.5em}
\end{figure}

To provide an intuitive illustration of the real-time performance, the exploration policy AION-e is directly evaluated on drones in IsaacSim, where two metrics are considered: CR and FCR. In each scene, the agent starts from the initial position and is allowed to explore the environment for 150 steps. AION-e is compared to ObjectNav baselines that integrate exploration and goal-reaching. To enforce exploration, the target object is set to a class that does not exist in the scene.
The average performance over 5 episodes is summarized in Table~\ref{tab:4}. 
Although ObjectNav baselines introduced in Table~\ref{tab:2} consider comprehensive skills, including exploration, goal-oriented navigation, and collision avoidance, they struggle to generalize to larger scenes and exhibit poor exploration performance. The exploration trajectories in \textit{Beechwood} are visualized in Fig.~\ref{fig:exp}, where goal-reaching policies tend to explore local areas when the target is invisible. In comparison, AION-e emphasizes exploration and safety during training, yielding a significantly higher FCR (\textbf{+7.2\%} in \textit{Chemistry}, \textbf{+20.6\%} in \textit{Beechwood}, and \textbf{+51.9\%} in \textit{Ihlen}) while maintaining a low CR.

\begin{table}[!t]
\centering
\small
\caption{Comparison of Exploration Performance in Isaacsim.}
\renewcommand{\arraystretch}{1.25}
\setlength{\tabcolsep}{2pt}
\begin{tabular}{ccccccc}
\toprule
\multirow{2}{*}{\textbf{Algorithm}} 
& \multicolumn{2}{c}{\textbf{Chemistry}} & \multicolumn{2}{c}{\textbf{Beechwood}} &
\multicolumn{2}{c}{\textbf{Ihlen}}\\ 
\cmidrule(lr){2-7}
& FCR$\uparrow$ & CR$\downarrow$& FCR$\uparrow$ & CR$\downarrow$& FCR$\uparrow$ & CR$\downarrow$ \\
\midrule
MJO &37.4& 8.0  & 39.2 & 14.7& 14.4 & 19.2\\
SSNet&34.3& 5.3  & 40.0 & 0.0& 13.9 & 1.5\\
AION-g &49.9 & 6.8  & 41.8 & \textbf{0.0}& 22.5 & 2.3\\
\rowcolor{gray!20} AION-e &\textbf{57.1}& \textbf{2.4}  & \textbf{62.4} & 0.3& \textbf{74.4} & \textbf{0.0}\\
\bottomrule
\end{tabular}
\label{tab:4}
\vspace{-1em}
\end{table}

\subsection{Evaluation of Dual-Policy ObjectNav}
The most effective strategy for aerial ObjectNav is to combine exploration and goal-reaching via policy switching. This section compares the dual-policy performance by pairing AION-e with different goal-reaching policies. To evaluate zero-shot generalization, IsaacSim experiments use four unseen target objects during training. Average results over 5 episodes are reported in Table~\ref{tab:5}, where dual-policy AION consistently attains the highest SR across all scenes. In most cases, CR is kept within a low range. 

\begin{table}[!t]
\centering
\small
\caption{ObjectNav Performance Across Scenes with Unseen Objects in Isaacsim.}
\renewcommand{\arraystretch}{1.25}
\setlength{\tabcolsep}{2pt}

\begin{tabular}{lcccccccccc}
\toprule
\multirow{2}{*}{\textbf{Algorithm}} 
& \multirow{2}{*}{\textbf{Objects}} 
& \multicolumn{2}{c}{\textbf{Chemistry}} & \multicolumn{2}{c}{\textbf{Beechwood}} &
\multicolumn{2}{c}{\textbf{Ihlen}}\\ 
\cmidrule(lr){3-8}
& & \textbf{SR}$\uparrow$ & \textbf{CR}$\downarrow$& \textbf{SR}$\uparrow$ & \textbf{CR}$\downarrow$& \textbf{SR}$\uparrow$ & \textbf{CR}$\downarrow$ \\
\midrule

\multirow{4}{*}{%
\begin{tabular}[l]{@{}l@{}}
AION-e\\
\& MJO
\end{tabular}}
& Sofa & 4/5 & 1.7  & 2/5 & \textbf{0.7}& 2/5 & 0.0\\
& Plant  & 2/5 & 1.2  & 4/5 & 1.2& 3/5 & \textbf{0.0}\\
& Laptop  & 0/5 & 1.1  & 3/5 & 7.2& 5/5 & 0.4\\
& Microwave  & 2/5 & 0.0  & 4/5 & 6.6& 2/5 & 2.3\\
\midrule
\multirow{4}{*}{%
\begin{tabular}[l]{@{}l@{}}
AION-e\\
\& SSNet
\end{tabular}}
& Sofa & 0/5 & 2.4  & 1/5 & 2.1& 2/5 & \textbf{0.0}\\
& Plant  & 3/5 & 1.4  & 1/5 & 0.7 & 2/5 & 1.8\\
& Laptop  & 0/5 & 0.7  & 3/5 & 2.2 & 2/5 & 0.0\\
& Microwave  & 1/5 & 4.4  & 4/5 & \textbf{1.0} & 3/5 & 1.3\\
\midrule
\multirow{4}{*}{%
\begin{tabular}[l]{@{}l@{}}
AION-e\\
\& AION-g
\end{tabular}}
& Sofa & \textbf{4/5} & \textbf{1.0} & \textbf{5/5} & 1.7& \textbf{4/5} & 1.5\\
& Plant & \textbf{5/5} & \textbf{0.1} & \textbf{5/5} & \textbf{0.4}& \textbf{4/5} & 0.2\\
& Laptop & \textbf{2/5} & \textbf{0.7}& \textbf{5/5} & \textbf{1.6} & \textbf{5/5} & \textbf{0.0}\\
& Microwave& \textbf{3/5} & \textbf{0.0} & \textbf{5/5} & 1.8 & \textbf{5/5} & \textbf{0.8}\\

\bottomrule
\end{tabular}
\label{tab:5}
\vspace{-1em}
\end{table}

\subsection{Failure Case Analysis}
For goal-reaching policy AION-g, when the target is in the front, it may occasionally prioritize goal-reaching while neglecting safety. That's the reason why CR doesn't reach zero.
In addition, since all algorithms heavily rely on the object detector to obtain feature $f_{\text{bbox}}$, misleading detection results constitute a primary cause of failure. In \textit{Chemistry} of Table~\ref{tab:5}, the low-light conditions severely degrade object detection and cause a low SR for the ``Laptop'' class due to the low visual contrast between the laptop and the background. Regarding exploration performance, both baselines and AION-g show limited capability in exploration due to the single-room training scenarios provided by the iTHOR benchmark. 
Although AION-e improves search efficiency through specialized training, ensuring full coverage remains unsolved for a purely vision-based approach.

\section{Conclusions}
In this paper, we present AION, a novel dual-policy RL framework for indoor ObjectNav with 3D locomotion and zero-shot generalization. The design of multi-modal inputs and reward functions is exploited to improve both exploration and goal-reaching performance. AION not only achieves SOTA results on iTHOR, but also validates its real-time performance under realistic drone dynamics. 
In summary, AION unlocks the potential of 3D ObjectNav in indoor environments and provides a solution for drone applications.

\bibliographystyle{IEEEtran}
\bibliography{IEEEabrv,reference}

@inproceedings{varghese2024yolov8,
  title={Yolov8: A novel object detection algorithm with enhanced performance and robustness},
  author={Varghese, Rejin and Sambath, M},
  booktitle={2024 International conference on advances in data engineering and intelligent computing systems (ADICS)},
  pages={1--6},
  year={2024},
  organization={IEEE}
}

@inproceedings{li2023blip,
  title={Blip-2: Bootstrapping language-image pre-training with frozen image encoders and large language models},
  author={Li, Junnan and Li, Dongxu and Savarese, Silvio and Hoi, Steven},
  booktitle={International conference on machine learning},
  pages={19730--19742},
  year={2023},
  organization={PMLR}
}

@inproceedings{procthor,
  author={Matt Deitke and Eli VanderBilt and Alvaro Herrasti and
          Luca Weihs and Jordi Salvador and Kiana Ehsani and
          Winson Han and Eric Kolve and Ali Farhadi and
          Aniruddha Kembhavi and Roozbeh Mottaghi},
  title={{ProcTHOR: Large-Scale Embodied AI Using Procedural Generation}},
  booktitle={NeurIPS},
  year={2022},
  note={Outstanding Paper Award}
}

@article{sun2025frontiernet,
  title={FrontierNet: Learning Visual Cues to Explore},
  author={Sun, Boyang and Chen, Hanzhi and Leutenegger, Stefan and Cadena, Cesar and Pollefeys, Marc and Blum, Hermann},
  journal={IEEE Robotics and Automation Letters},
  year={2025},
}

@inproceedings{zhou2023esc,
  title={Esc: Exploration with soft commonsense constraints for zero-shot object navigation},
  author={Zhou, Kaiwen and Zheng, Kaizhi and Pryor, Connor and Shen, Yilin and Jin, Hongxia and Getoor, Lise and Wang, Xin Eric},
  booktitle={International Conference on Machine Learning},
  pages={42829--42842},
  year={2023},
}

@article{liu2025indooruav,
  title={IndoorUAV: Benchmarking Vision-Language UAV Navigation in Continuous Indoor Environments},
  author={Liu, Xu and Liu, Yu and Qiu, Hanshuo and Qirong, Yang and Lian, Zhouhui},
  journal={arXiv preprint arXiv:2512.19024},
  year={2025}
}

@article{wang2025uav,
  title={UAV-Flow Colosseo: A Real-World Benchmark for Flying-on-a-Word UAV Imitation Learning},
  author={Wang, Xiangyu and Yang, Donglin and Liao, Yue and Zheng, Wenhao and Dai, Bin and Li, Hongsheng and Liu, Si and others},
  journal={arXiv preprint arXiv:2505.15725},
  year={2025}
}

@inproceedings{dang2023search,
  title={Search for or navigate to? dual adaptive thinking for object navigation},
  author={Dang, Ronghao and Wang, Liuyi and He, Zongtao and Su, Shuai and Tang, Jiagui and Liu, Chengju and Chen, Qijun},
  booktitle={Proceedings of the IEEE/CVF International Conference on Computer Vision},
  pages={8250--8259},
  year={2023}
}

@inproceedings{sun2024prioritized,
  title={Prioritized semantic learning for zero-shot instance navigation},
  author={Sun, Xinyu and Liu, Lizhao and Zhi, Hongyan and Qiu, Ronghe and Liang, Junwei},
  booktitle={European Conference on Computer Vision},
  pages={161--178},
  year={2024},
}

@article{peng2025logoplanner,
  title={LoGoPlanner: Localization Grounded Navigation Policy with Metric-aware Visual Geometry},
  author={Peng, Jiaqi and Cai, Wenzhe and Yang, Yuqiang and Wang, Tai and Shen, Yuan and Pang, Jiangmiao},
  journal={arXiv preprint arXiv:2512.19629},
  year={2025}
}

@inproceedings{DBLP:conf/iclr/PuigUSCYPDCHMVG24,
  author       = {Xavier Puig and
                  Eric Undersander and
                  Andrew Szot and
                  Mikael Dallaire Cote and
                  Tsung{-}Yen Yang and
                  Ruslan Partsey and
                  Ruta Desai and
                  Alexander Clegg and
                  Michal Hlavac and
                  So Yeon Min and
                  Vladimir Vondrus and
                  Th{\'{e}}ophile Gervet and
                  Vincent{-}Pierre Berges and
                  John M. Turner and
                  Oleksandr Maksymets and
                  Zsolt Kira and
                  Mrinal Kalakrishnan and
                  Jitendra Malik and
                  Devendra Singh Chaplot and
                  Unnat Jain and
                  Dhruv Batra and
                  Akshara Rai and
                  Roozbeh Mottaghi},
  title        = {Habitat 3.0: {A} Co-Habitat for Humans, Avatars, and Robots},
  booktitle    = {The Twelfth International Conference on Learning Representations, {ICLR} 2024, Vienna, Austria, May 7-11, 2024},
  year         = {2024},
}

@inproceedings{cai2024bridging,
  title={Bridging zero-shot object navigation and foundation models through pixel-guided navigation skill},
  author={Cai, Wenzhe and Huang, Siyuan and Cheng, Guangran and Long, Yuxing and Gao, Peng and Sun, Changyin and Dong, Hao},
  booktitle={2024 IEEE International Conference on Robotics and Automation (ICRA)},
  pages={5228--5234},
  year={2024},
}

@article{zhang2025apexnav,
  title={ApexNav: An Adaptive Exploration Strategy for Zero-Shot Object Navigation with Target-centric Semantic Fusion},
  author={Zhang, Mingjie and Du, Yuheng and Wu, Chengkai and Zhou, Jinni and Qi, Zhenchao and Ma, Jun and Zhou, Boyu},
  journal={arXiv preprint arXiv:2504.14478},
  year={2025}
}

@inproceedings{yokoyama2024vlfm,
  title={Vlfm: Vision-language frontier maps for zero-shot semantic navigation},
  author={Yokoyama, Naoki and Ha, Sehoon and Batra, Dhruv and Wang, Jiuguang and Bucher, Bernadette},
  booktitle={2024 IEEE International Conference on Robotics and Automation (ICRA)},
  pages={42--48},
  year={2024},
}

@article{gervet2023navigating,
  title={Navigating to objects in the real world},
  author={Gervet, Theophile and Chintala, Soumith and Batra, Dhruv and Malik, Jitendra and Chaplot, Devendra Singh},
  journal={Science Robotics},
  volume={8},
  number={79},
  pages={eadf6991},
  year={2023},
  publisher={American Association for the Advancement of Science}
}

@article{oquab2023dinov2,
  title={Dinov2: Learning robust visual features without supervision},
  author={Oquab, Maxime and Darcet, Timoth{\'e}e and Moutakanni, Th{\'e}o and Vo, Huy and Szafraniec, Marc and Khalidov, Vasil and Fernandez, Pierre and Haziza, Daniel and Massa, Francisco and El-Nouby, Alaaeldin and others},
  journal={arXiv preprint arXiv:2304.07193},
  year={2023}
}

@inproceedings{mjo,
  title={Learning hierarchical relationships for object-goal navigation},
  author={Pal, Anwesan and Qiu, Yiding and Christensen, Henrik},
  booktitle={Conference on Robot Learning},
  pages={517--528},
  year={2021},
}

@inproceedings{clip,
  title={Learning transferable visual models from natural language supervision},
  author={Radford, Alec and Kim, Jong Wook and Hallacy, Chris and Ramesh, Aditya and Goh, Gabriel and Agarwal, Sandhini and Sastry, Girish and Askell, Amanda and Mishkin, Pamela and Clark, Jack and others},
  booktitle={International conference on machine learning},
  pages={8748--8763},
  year={2021},
}

@article{tdanet,
  title={Tdanet: Target-directed attention network for object-goal visual navigation with zero-shot ability},
  author={Lian, Shiwei and Zhang, Feitian},
  journal={IEEE Robotics and Automation Letters},
  year={2024},
}

@INPROCEEDINGS{zson,
  author={Zhao, Qianfan and Zhang, Lu and He, Bin and Qiao, Hong and Liu, Zhiyong},
  booktitle={2023 IEEE International Conference on Robotics and Automation (ICRA)}, 
  title={Zero-Shot Object Goal Visual Navigation}, 
  year={2023},
  pages={2025-2031}
}

@article{sc,
  title={Visual semantic navigation using scene priors},
  author={Yang, Wei and Wang, Xiaolong and Farhadi, Ali and Gupta, Abhinav and Mottaghi, Roozbeh},
  journal={arXiv preprint arXiv:1810.06543},
  year={2018}
}

@inproceedings{savn,
  title={Learning to learn how to learn: Self-adaptive visual navigation using meta-learning},
  author={Wortsman, Mitchell and Ehsani, Kiana and Rastegari, Mohammad and Farhadi, Ali and Mottaghi, Roozbeh},
  booktitle={Proceedings of the IEEE/CVF conference on computer vision and pattern recognition},
  pages={6750--6759},
  year={2019}
}

@article{ai2thor,
  title={Ai2-thor: An interactive 3d environment for visual ai},
  author={Kolve, Eric and Mottaghi, Roozbeh and Han, Winson and VanderBilt, Eli and Weihs, Luca and Herrasti, Alvaro and Deitke, Matt and Ehsani, Kiana and Gordon, Daniel and Zhu, Yuke and others},
  journal={arXiv preprint arXiv:1712.05474},
  year={2017}
}

@inproceedings{glove,
  title={Glove: Global vectors for word representation},
  author={Pennington, Jeffrey and Socher, Richard and Manning, Christopher D},
  booktitle={Proceedings of the 2014 conference on empirical methods in natural language processing (EMNLP)},
  pages={1532--1543},
  year={2014}
}

@inproceedings{Pegasus,
  author={Jacinto, Marcelo and Pinto, João and Patrikar, Jay and Keller, John and Cunha, Rita and Scherer, Sebastian and Pascoal, António},
  booktitle={2024 International Conference on Unmanned Aircraft Systems (ICUAS)}, 
  title={Pegasus Simulator: An Isaac Sim Framework for Multiple Aerial Vehicles Simulation}, 
  year={2024},
  pages={917-922}
}

@inproceedings{stubborn,
  title={Stubborn: A strong baseline for indoor object navigation},
  author={Luo, Haokuan and Yue, Albert and Hong, Zhang-Wei and Agrawal, Pulkit},
  booktitle={2022 IEEE/RSJ International Conference on Intelligent Robots and Systems (IROS)},
  pages={3287--3293},
  year={2022}
}

@article{chaplot2020object,
  title={Object goal navigation using goal-oriented semantic exploration},
  author={Chaplot, Devendra Singh and Gandhi, Dhiraj Prakashchand and Gupta, Abhinav and Salakhutdinov, Russ R},
  journal={Advances in Neural Information Processing Systems},
  volume={33},
  pages={4247--4258},
  year={2020}
}

@inproceedings{du2023object,
  title={Object-goal visual navigation via effective exploration of relations among historical navigation states},
  author={Du, Heming and Li, Lincheng and Huang, Zi and Yu, Xin},
  booktitle={Proceedings of the IEEE/CVF Conference on Computer Vision and Pattern Recognition},
  pages={2563--2573},
  year={2023}
}

@inproceedings{yu2023l3mvn,
  title={L3mvn: Leveraging large language models for visual target navigation},
  author={Yu, Bangguo and Kasaei, Hamidreza and Cao, Ming},
  booktitle={2023 IEEE/RSJ International Conference on Intelligent Robots and Systems (IROS)},
  pages={3554--3560},
  year={2023}
}

@inproceedings{omnigibson,
  title={Behavior-1k: A benchmark for embodied ai with 1,000 everyday activities and realistic simulation},
  author={Li, Chengshu and Zhang, Ruohan and Wong, Josiah and Gokmen, Cem and Srivastava, Sanjana and Mart{\'\i}n-Mart{\'\i}n, Roberto and Wang, Chen and Levine, Gabrael and Lingelbach, Michael and Sun, Jiankai and others},
  booktitle={Conference on Robot Learning},
  pages={80--93},
  year={2023}
}

@software{NVIDIA_Isaac_Sim,
author = {{NVIDIA}},
license = {Apache-2.0},
title = {{Isaac Sim}},
url = {https://github.com/isaac-sim/IsaacSim},
version = {5.0.0}
}

@inproceedings{lee2025citynav,
  title={CityNav: A Large-Scale Dataset for Real-World Aerial Navigation},
  author={Lee, Jungdae and Miyanishi, Taiki and Kurita, Shuhei and Sakamoto, Koya and Azuma, Daichi and Matsuo, Yutaka and Inoue, Nakamasa},
  booktitle={Proceedings of the IEEE/CVF International Conference on Computer Vision},
  pages={5912--5922},
  year={2025}
}

@inproceedings{xiao2025uav,
  title={Uav-on: A benchmark for open-world object goal navigation with aerial agents},
  author={Xiao, Jianqiang and Sun, Yuexuan and Shao, Yixin and Gan, Boxi and Liu, Rongqiang and Wu, Yanjin and Guan, Weili and Deng, Xiang},
  booktitle={Proceedings of the 33rd ACM International Conference on Multimedia},
  pages={13023--13029},
  year={2025}
}

@inproceedings{liu2023aerialvln,
  title={Aerialvln: Vision-and-language navigation for uavs},
  author={Liu, Shubo and Zhang, Hongsheng and Qi, Yuankai and Wang, Peng and Zhang, Yanning and Wu, Qi},
  booktitle={Proceedings of the IEEE/CVF International Conference on Computer Vision},
  pages={15384--15394},
  year={2023}
}

\end{document}